# *YOLO26: A Comprehensive Architecture Overview and Key Improvements*

A PREPRINT

**Priyanto Hidayatullah[a*], Refdinal Tubagus[b],**
[a] Computer Engineering and Informatics Department, Politeknik Negeri Bandung, Kab. Bandung Barat, Indonesia
[b] Stunning Vision AI, Kota Cimahi, Indonesia
priyanto@polban.ac.id[a], refdinal@stunningvisionai.com[b]

**Corresponding Author*

February 16, 2026

**ABSTRACT**

You Only Look Once (YOLO) has been the prominent model for computer vision in deep learning for a decade. This study explores the novel aspects of YOLO26, the most recent version in the YOLO series. The elimination of Distribution Focal Loss (DFL), implementation of End-to-End NMS-Free Inference, introduction of ProgLoss + Small-Target-Aware Label Assignment (STAL), and use of the MuSGD optimizer are the primary enhancements designed to improve inference speed, which is claimed to achieve a 43% boost in CPU mode. This is designed to allow YOLO26 to attain real-time performance on edge devices or those without GPUs. Additionally, YOLO26 offers improvements in many computer vision tasks, including instance segmentation, pose estimation, and oriented bounding box (OBB) decoding. We aim for this effort to provide more value than just consolidating information already included in the existing technical documentation. Therefore, we performed a rigorous architectural investigation into YOLO26, mostly using the source code available in its GitHub repository and its official documentation. The authentic and detailed operational mechanisms of YOLO26 are inside the source code, which is seldom extracted by others. The YOLO26 architectural diagram is shown as the outcome of the investigation. This study is, to our knowledge, the first one presenting the CNN-based YOLO26 architecture, which is the core of YOLO26. Our objective is to provide a precise architectural comprehension of YOLO26 for researchers and developers aspiring to enhance the YOLO model, ensuring it remains the leading deep learning model in computer vision.

## 1 Introduction

The Ultralytics' version of YOLO is the most awaited version of YOLO. It has several advantages. [1]:

a. It includes many features that tackle computer vision challenges: object detection, instance segmentation, image classification, pose estimation, detection of oriented objects, and object tracking. It represents a comprehensive bundle.
b. It is straightforward to execute and enhance. YAML-based architectural definitions are readily adjustable.
c. It has robust deployment assistance. It can be exported to ONNX, TensorRT, CoreML, and OpenVINO.
d. It facilitates quantization, which enhances speed.
e. It offers a balance between speed and accuracy. Similar to past YOLO versions, Ultralytics YOLO upholds excellent accuracy while maintaining real-time inference speed. Particularly in YOLO26, enhancing speed for edge devices is the primary focus.

YOLO26 is a comprehensive model for addressing various computer vision problems as mentioned above. Nonetheless, YOLO's functionalities have been completely implemented in YOLOv8 [2]. From this point of view, this version has no substantial enhancements. However, YOLO26 improves those additional functionalities, which are elaborated in the fifth section of this paper.

The designation of the YOLO26 name was interestingly atypical. Rather than continuing the previous version numbering, YOLOv13 [3], it jumped to YOLO26. Some might think that it is the 2026 version of YOLO, because it was released on 14 January 2026. However, a question may arise: what if other institutions or a group of people choose to launch a new version of YOLO this year as well?

Humans are inherently visual beings. They work better when something is visualized in pictures. Verbal and numerical expressions are frequently inadequate. The absence of a YOLO26 architectural diagram creates a substantial void for researchers and developers in comprehending and advancing the model to a higher standard. This is essential for YOLO's ongoing advancement and its capability to compete with other object detection models, including RF-DETR [4], and RT-DETRv3 [5] transformer-based models. To maintain YOLO's prominence in object detection models and computer vision as a whole, we should encourage the community in developing YOLO model. The provision of the architectural diagram and its explanation are key factors for the community's advancement in enhancing the model. Architectural improvements in one version of YOLO may influence another, which is now occurring: YOLO26 included some improvements from YOLOv10 [6].

To obtain an accurate architectural diagram, one cannot just rely on the technical documentation for YOLO26 [1]. For example, we will not know in which block the Distribution Focal Loss (DFL) is located within the architecture just by reading the documentation. To ascertain the location of each architectural modification in YOLO26 and how it works, it is necessary to look through the source code of YOLO26 directly [7]. To our knowledge, this article is the first paper to provide an overall YOLO26 architecture diagram. In addition, we also elaborate on YOLO26 key improvements. We hope that this work will contribute to the development of an improved YOLO model.

Based on our in-depth exploration, YOLO26 has several architectural improvements compared to its earlier version. Nonetheless, the general design remains unchanged from the last version, which is a single-stage and end-to-end object detector. The improvements seek to enhance efficiency, stabilize training, and remove dependence on non-maximum suppression during the prediction phase. Aligned with its launch slogan, "Built End-to-End. Built for the Edge," YOLO26 seeks to enhance performance on edge devices while simultaneously increasing accuracy relative to its predecessor.

## 2 Evolution of YOLO Models

Table 1 shows YOLO model evolution. The YOLO model has changed a lot over time, and each new version has made improvements to the architecture. This shows how quickly real-time object detection is improving. Three versions were released over the course of four years. However, since YOLOv4 was released in 2020, there have been 11 versions from 2020 to 2026. This indicates that YOLO is gaining popularity and advancing rapidly.

Table 1: Evolution of YOLO Models

| No | Version | Release (Month–Year) | Author(s) | Key Architectural Innovations |
|---|---|---|---|---|
| 1 | **YOLOv1** [8] | Jun 2016 | Joseph Redmon, Santosh Divvala, Ross Girshick, Ali Farhadi | Unified one-stage detection; grid-based prediction; direct bounding-box regression |
| 2 | **YOLOv2 (YOLO9000)** [9] | Dec 2016 | Joseph Redmon, Ali Farhadi | Anchor boxes; Batch Normalization; dimension clustering; passthrough layer |
| 3 | **YOLOv3** [10] | Apr 2018 | Joseph Redmon, Ali Farhadi | Darknet-53 backbone; residual blocks; multi-scale feature prediction |
| 4 | **YOLOv4** [11] | Apr 2020 | Alexey Bochkovskiy, Chien-Yao Wang, Hong-Yuan Mark Liao | CSPDarknet53; PANet; SPP; bag-of-freebies & bag-of-specials integration |
| 5 | **YOLOv5** [12] | Jun 2020 | Glenn Jocher (Ultralytics) | PyTorch implementation; CSP bottleneck variants; modular design; auto-anchor |
| 6 | **YOLOv6** [13] | Jun 2022 | Meituan Vision | Anchor-free design; EfficientRep backbone; decoupled head design |
| 7 | **YOLOv7** [14] | Jul 2022 | Chien-Yao Wang et al. | E-ELAN; re-parameterized convolution; enhanced feature aggregation |
| 8 | **YOLOv8** [2] | Jan 2023 | Ultralytics | Fully anchor-free head; decoupled classification/regression; Distribution Focal Loss (DFL) integration |
| 9 | **YOLOv9** [15] | Apr 2024 | Chien-Yao Wang et al. | Programmable Gradient Information (PGI) mechanisms; GELAN backbone variants |
| 10 | **YOLOv10** [6] | May 2024 | Guangxing Han et al. | NMS-free training; dual label assignment (one-to-many & one-to-one) |
| 11 | **YOLOv11** [16] | Jul 2024 | Ultralytics | C3k2 blocks; kernel-optimized modules; performance scaling enhancements |
| 12 | **YOLOv12** [17] | Feb 2025 | Yunjie Tian, Qixiang Ye, David Doermann | Attention-centric design; Area Attention & R-ELAN; optimized attention for real-time inference |
| 13 | **YOLOv13** [3] | Jun 2025 | Mengqi Lei, Siqi Li, Yihong Wu, Han Hu, You Zhou, Xinhu Zheng, Guiguang Ding, Shaoyi Du, Zongze Wu, Yue Gao | Hypergraph-based Adaptive Correlation Enhancement (HyperACE); Full-Pipeline Aggregation-and-Distribution (FullPAD); depthwise separable blocks |
| 14 | **YOLO26** [1] | Jan 2026 | Ultralytics | SPPF shortcut; PSABlock in final C3k2; removes DFL; dual assignment NMS-free head; Top-K score-based inference; MuSGD Optimizer; ProgLoss + STAL |

Beginning with YOLOv1, which transformed object recognition through a one-stage grid-based approach and direct bounding box regression, the YOLO model has evolved significantly. Subsequent versions have incorporated enhancements such as anchor boxes, batch normalization, and multi-scale prediction, building upon the foundation laid by its initial versions. Subsequent innovations, including the development of more profound and efficient architectures (Darknet-53, CSPDarknet, EfficientRep), refined feature aggregation methods (PANet, SPP, ELAN, GELAN), and distinctive anchor-free head designs, illustrate YOLO's commitment to improving both speed and accuracy.

The change from YOLOv9 to YOLOv13 is a step toward a more flexible design. This includes features like Programmable Gradient Information, attention-based structures, and hypergraphs for correlation modeling. YOLO26 advanced these concepts by implementing a non-NMS approach that incorporates dual label assignment. This innovation simplifies the process by eliminating Distribution Focal Loss to enhance end-to-end efficiency, alongside utilizing the MuSGD Optimizer and integrating ProgLoss with STAL. The progression of YOLO demonstrates improvements in the effectiveness of object detection, while also enhancing its capability to a range of contemporary computer vision tasks, including instance segmentation, pose estimation, and oriented bounding boxes.

## 3 YOLO26 Architecture

This architectural diagram is derived from the YOLO26 architecture file, yolo26.yaml located in the ultralytics/cfg/models/26 directory and tasks.py located in the ultralytics/nn directory. The code was obtained from Ultralytics GitHub repository, using the latest version (release 8.4.14). The architecture diagram is depicted in Figure 1. We conducted input and output tensor tracing to validate the architectural diagram.

Like YOLOv8 and YOLO11, the YOLO26 variant is defined by three parameters. The parameters include depth_multiple, width_multiple, and max_channels. Depth_multiple specifies the quantity of bottleneck blocks in the C3k2 module and the number of PSA blocks in the C2PSA module. Simultaneously, width_multiple and max_channels determine the quantity of output channels for each block. YOLO26 requires a three-channel image as input. This input will go to the backbone, neck region, and ultimately the head.

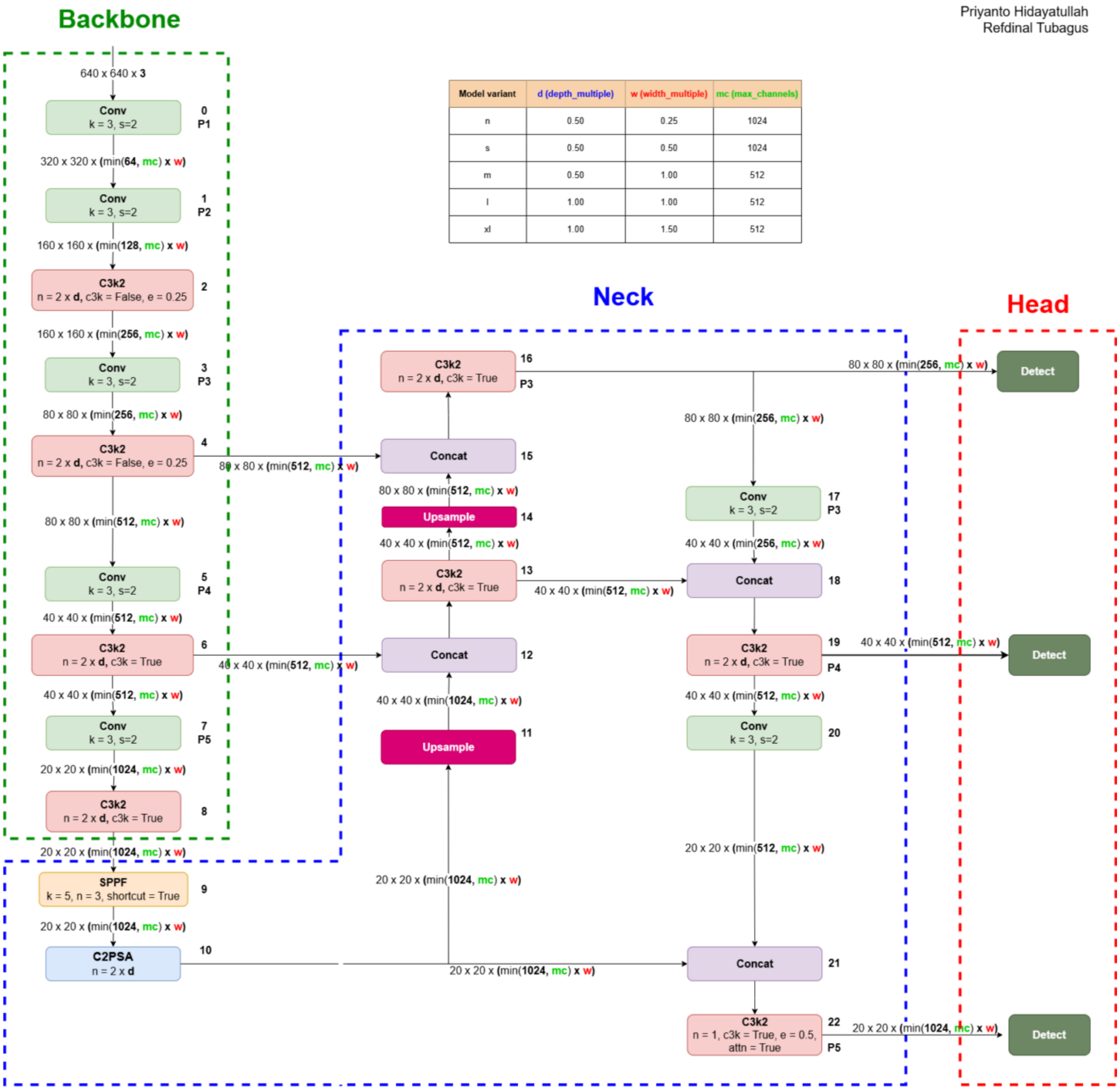

| Model variant | d (depth_multiple) | w (width_multiple) | mc (max_channels) |
|---|---|---|---|
| n | 0.50 | 0.25 | 1024 |
| s | 0.50 | 0.50 | 1024 |
| m | 0.50 | 1.00 | 512 |
| l | 1.00 | 1.00 | 512 |
| xl | 1.00 | 1.50 | 512 |

Figure 1: YOLO26 Architecture Diagram

The backbone begins with two convolutional blocks using a kernel size of 3 and a stride of 2. The use of a stride of 2 decreases the spatial resolution of the feature maps. For every convolutional block, the output feature map spatial resolution is reduced to half of the input resolution. The subsequent component is the C3k2 block, used for generating features with a significant degree of abstraction. This block has many parameters, namely n, c3k, and e. Following that, inside the backbone, there are several more convolutional blocks, namely blocks 3, 5, and 7. Additionally, there are numerous additional C3k2 blocks, namely blocks 4, 6, and 8. These three blocks are linked to the neck.

The first component in the neck is SPPF (Spatial Pyramid Pooling Fast), enabling the model to extract features at several levels of abstraction by pooling at different sizes. The SPPF block in YOLO26 has been enhanced with the incorporation of a shortcut. This shortcut allows the immediate incorporation of input into the output, hence enhancing information flow and augmenting feature representation. The C2PSA block employs a self-attention method to enhance efficiency via the integration of global modeling capabilities. Subsequently, many upsample and concat blocks are present. Upsampling is used to enhance the resolution of the feature map by the closest neighbor upsampling method. Concat is used for merging feature maps. During this process, the resolution remains constant, but the number of channels increases. The neck comprises two convolutional blocks and four C3k2 blocks, three of which are linked to the head. In the final C3k2 block, a new component has been incorporated: an attention block.

YOLO26 comprises three heads. The initial head, linked to C3k2 block number 16, is used for the detection of small objects. The secondary head, linked to C3k2 block number 19, is used for the detection of medium-sized objects. The third head, associated with block C3k2 number 22, is used for the detection of big objects. It is essential to notice that in every YOLO version, the dimensions of objects are proportional to the size of the image or video frame.

# 4 YOLO26 Key Improvements Compared to Previous YOLO versions

The YOLO26 architecture design is very similar to the YOLO11 architecture [18]. Both have many similar architectural components and both were made by the same company, Ultralytics. Still, YOLO26 introduces new ideas compared to previous YOLO versions. The following are the improvements and modifications that have been made since YOLO11.

Structural modifications have been implemented in the Spatial Pyramid Pooling Fast (SPPF) block. YOLO26 incorporates a shortcut connection inside the SPPF. This connection enhances gradient transmission across feature sets and stabilizes optimization in the context of high-level semantic representations.

An additional modification is seen in the last C3k2 block before the detect block. In YOLO26, the repeat parameter n is set at 1. Augmenting the number of repetitions just escalates the computing overhead without enhancing accuracy. To address this, YOLO26 incorporates an attention mechanism inside the PSABlock module. This approach enhances global context modeling while minimizing the rise in parameters and latency.

The head part of YOLO26 still has three detection blocks, each of which specialize for detecting small, medium, and big sized object. The most notable difference is that YOLO26 has eliminated the usage of Distributed Focal Loss (DFL) in the detect block. It is replaced by box regression, which is managed by explicitly predicting coordinates. This streamlines both the training and inference process.

In previous YOLO models, DFL was used to improve bounding box regression by predicting the distribution of possible bounding box locations, rather than a single value. DFL adds extra computation and a fixed regression range, making it difficult for the model to learn one-to-one object assignments and increasing its dependence on Non-Maximum Suppression (NMS). In YOLO26, DFL is removed. Instead of relying on distribution-based outputs, the model now learns to predict accurate bounding box coordinates in a way that supports fewer but more confident detections [1].

The Head part employs dual assignment for NMS Free Training, which is inspired by YOLOv10 [1], [6]. During the training process, both one-to-many and one-to-one assignments are used. The objective is for the backbone and neck to have comprehensive guidance from the one-to-many assignment. During inference, the one-to-many head is eliminated, and just the one-to-one head is used to provide predictions.

For more details, YOLO26 eliminates the necessity for post-processing procedures, which is Non-Maximum Suppression (NMS). When there are multiple predictions in a single frame, NMS often causes problems. YOLO26 does not generate multiple overlapping predictions followed by a filtering process. Instead, it transmits the final detections directly from the network. This is referred to as End-to-End NMS-Free Inference. YOLO26 employs two detection heads in its training process, each utilizing the same foundational model but pursuing distinct objectives. The one-to-one head associates each object with a distinct and specific prediction, which is crucial for the end-to-end NMS-Free architecture. Simultaneously, the one-to-many head is utilized exclusively during the training phase. This head enables the connection of multiple predictions to a single object, thereby enhancing the density of supervision. This enhanced learning signal contributes to maintaining training stability and increasing accuracy, particularly during the initial stages.

YOLO26 uses Progressive Loss Balancing to adjust the weight of the learning signal over time. ProgLoss works by dynamically shifting the contribution of each head to the total loss throughout the training process. In the early stages, greater weight is given to the one-to-many head to stabilize learning and improve recall. As training progresses, this balance gradually shifts towards the one-to-one head, bringing the training process more in line with inference behavior. The result is smoother convergence, fewer unstable training runs, and more consistent final performance. In addition, YOLO26 improves upon the existing label assignment method, Task Alignment Learning (TAL), which is designed to better align classification and localization during training. However, TAL often ignores very small objects. Therefore, YOLO26 introduces Small-Target-Aware Label Assignment (STAL). STAL modifies the assignment process so that small objects are not ignored during training. Specifically, this method sets a minimum of four anchors for objects smaller than 8 x 8 pixels for a 640 x 640 pixels input image. Thus, even very small objects consistently contribute to training loss.

To support a more stable and predictable training process, Ultralytics YOLO26 also introduces a new optimizer called MuSGD. MuSGD builds on this familiar foundation by incorporating optimization ideas inspired by Muon [19], a method used in training large language models. YOLO26 uses a hybrid update strategy. Some parameters are updated using a combination of Muon-style updates and Stochastic Gradient Descent (SGD), while others remain purely SGD. The result is a smoother optimization process, faster convergence, and more predictable training behavior across model sizes.

During the inference phase, the selection of detection outcomes is determined not by comparing IoU across bounding boxes as in Non-Maximum Suppression, but via a score-based ranking methodology. The model executes Top-K selection based on classification scores, specifically identifying a set of predictions with the highest global confidence values. This process does not involve the computation of Intersection over Union (IoU) or non-maximum suppression between bounding boxes.

With the above improvements, YOLO26 seeks to increase its accuracy beyond previous versions of YOLO. In addition, the developers of YOLO26 claimed that YOLO26 can increase speed in CPU mode by up to 43%. This is very useful for implementing YOLO26 on edge devices or devices without GPUs [1].

## 5 YOLO26 Computer Vision Specific Tasks Improvement

Ultralytics YOLO supports many Computer Vision tasks, as mentioned in the beginning of the introduction section. YOLO26 improves its performance for tackling those tasks. We explore the improvement in detail.

a. Object Detection
   Object detection is the main feature supported by YOLO26. This task involves identifying objects in images or video frames and drawing bounding boxes around them. Detected objects are then classified into various categories based on their features. Various developments and modifications to the YOLO26 architecture directly affect YOLO26's performance in object detection tasks.

b. Instance Segmentation
   Segmentation takes object detection a step further by generating pixel-level masks for each object. This level of precision is useful for applications such as medical imaging, agricultural analysis, and manufacturing quality control. YOLO26 introduces semantic segmentation loss to improve model convergence and an enhanced proto module that utilizes multi-scale information for superior mask quality [1].

c. Image Classification
   Image classification is a task that involves assigning an entire image to one of a predefined set of classes. The output of an image classifier is a single class label along with a confidence score. Image classification is useful when you only need to determine which class an image belongs to, without identifying the location or exact shape of objects within the image [20].

d. Pose Estimation
   Pose Estimation detects specific keypoints in images or video frames to track movement or estimate pose. These keypoints can represent human joints, facial features, or other important points. YOLO26 integrates Residual Log-Likelihood Estimation (RLE) for more accurate keypoint localization and optimizes the decoding process to improve inference speed [1].

e. Oriented Bounding Box (OBB)
   Oriented Bounding Box (OBB) detection improves traditional object detection by adding orientation angles to better locate rotating objects. YOLO26 introduces specialized angle loss to improve detection accuracy for square-shaped objects and optimizes OBB decoding to address boundary discontinuity issues [1].

## 6 Performance Comparison

Figure 2 shows a comparison between YOLO26 and previous versions of YOLO, which is based on YOLO26's technical documentation [1]. The mean average precision (mAP) calculations were performed on the COCO dataset, while speed was measured using an NVIDIA GPU T4. YOLO26 has the highest accuracy compared to previous versions in all variants. However, in terms of speed, only the s and m variants are the fastest, and their speed is equivalent to YOLO11 [1].

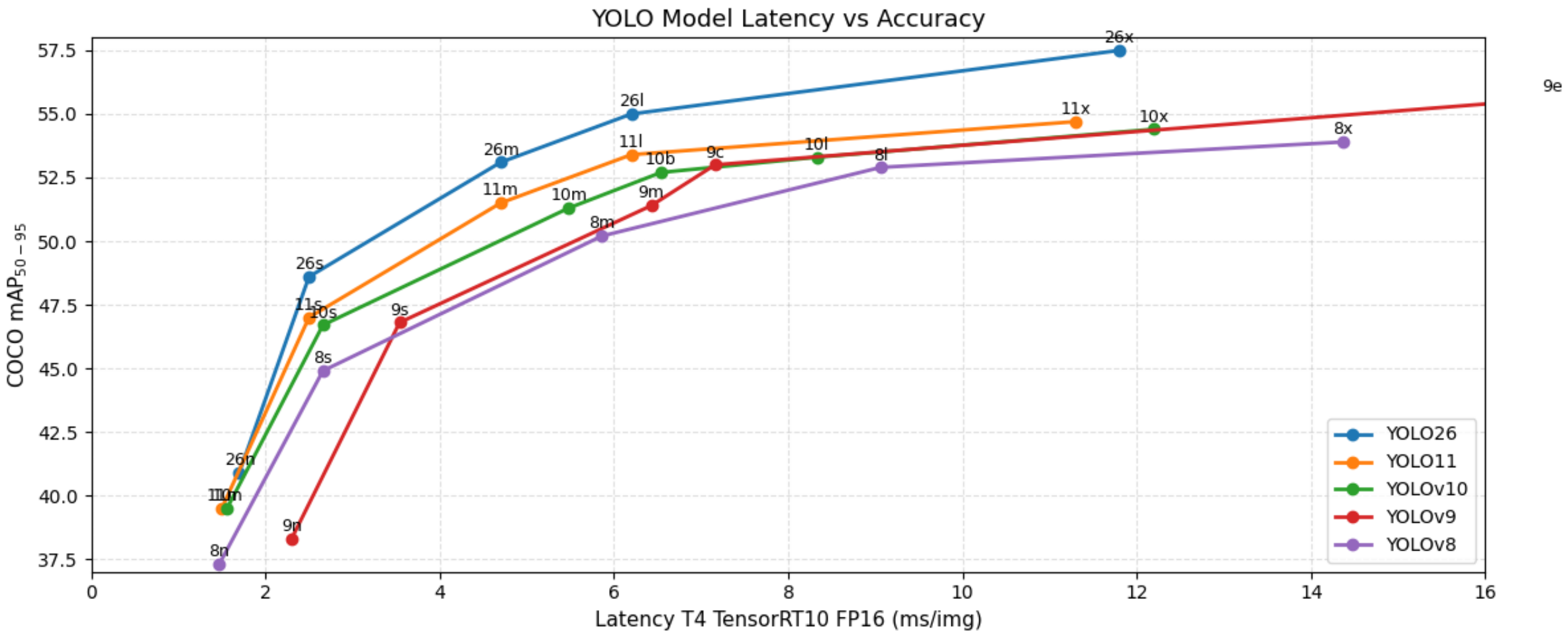

Figure 2: YOLO Performance Comparison [1], [16]

## 7 Conclusion

YOLO26 introduces several improvements in architecture and training mechanisms compared to previous versions of YOLO. Based on the claimed results of the YOLO models' performance comparison, these improvements have resulted in better performance. From an architectural perspective, YOLO26 is an improved version of its predecessor, but not a disruptive redesign. The results of the deep exploration provided in this paper conclude that YOLO26 refines many of its architectural components to be more accurate (especially for detecting small objects), simplifies inference, and reduces computational cost, which is suitable for edge devices.

## References

[1] Ultralytics, “Ultralytics YOLO26.” Accessed: Jan. 31, 2026. [Online]. Available: https://docs.ultralytics.com/models/yolo26/

[2] Glenn Jocher, Ayush Chaurasia, and Jing Qiu, *YOLOv8*. (Jan. 23, 2023). Ultralytics. [Online]. Available: https://github.com/ultralytics/ultralytics

[3] M. Lei *et al.*, “YOLOv13: Real-Time Object Detection with Hypergraph-Enhanced Adaptive Visual Perception,” 2025, *arXiv*. doi: 10.48550/ARXIV.2506.17733.

[4] I. Robinson, P. Robicheaux, M. Popov, D. Ramanan, and N. Peri, “RF-DETR: Neural Architecture Search for Real-Time Detection Transformers,” 2025, *arXiv*. doi: 10.48550/ARXIV.2511.09554.

[5] S. Wang, C. Xia, F. Lv, and Y. Shi, “RT-DETRv3: Real-Time End-to-End Object Detection with Hierarchical Dense Positive Supervision,” in *2025 IEEE/CVF Winter Conference on Applications of Computer Vision (WACV)*, Tucson, AZ, USA: IEEE, Feb. 2025, pp. 1628–1636. doi: 10.1109/WACV61041.2025.00166.

[6] A. Wang *et al.*, “YOLOv10: Real-Time End-to-End Object Detection,” 2024, *arXiv*. doi: 10.48550/ARXIV.2405.14458.

[7] G. Jocher, J. Qiu, and A. Chaurasia, *Ultralytics YOLO*. (Jan. 2023). Python. Accessed: Aug. 22, 2025. [Online]. Available: https://github.com/ultralytics/ultralytics

[8] J. Redmon, S. Divvala, R. Girshick, and A. Farhadi, “You Only Look Once: Unified, Real-Time Object Detection,” in *2016 IEEE Conference on Computer Vision and Pattern Recognition (CVPR)*, Las Vegas, NV, USA: IEEE, Jun. 2016, pp. 779–788. doi: 10.1109/CVPR.2016.91.

[9] J. Redmon and A. Farhadi, “YOLO9000: Better, Faster, Stronger,” in *2017 IEEE Conference on Computer Vision and Pattern Recognition (CVPR)*, IEEE, Jul. 2017. doi: 10.1109/CVPR.2017.690.

[10] J. Redmon and A. Farhadi, “YOLOv3: An Incremental Improvement,” *arXiv:1804.02767 [cs]*, Apr. 2018, Accessed: Jun. 02, 2021. [Online]. Available: http://arxiv.org/abs/1804.02767

[11] A. Bochkovskiy, C.-Y. Wang, and H.-Y. M. Liao, “YOLOv4: Optimal Speed and Accuracy of Object

Detection," *arXiv:2004.10934 [cs, eess]*, Apr. 2020, Accessed: Jun. 02, 2021. [Online]. Available: http://arxiv.org/abs/2004.10934
[12] "Releases · ultralytics/yolov5 · GitHub." Accessed: Jan. 31, 2026. [Online]. Available: https://github.com/ultralytics/yolov5/releases
[13] C. Li *et al.*, "YOLOv6: A Single-Stage Object Detection Framework for Industrial Applications," Sep. 07, 2022, *arXiv*: arXiv:2209.02976. doi: 10.48550/arXiv.2209.02976.
[14] C.-Y. Wang, A. Bochkovskiy, and H.-Y. M. Liao, "YOLOv7: Trainable bag-of-freebies sets new state-of-the-art for real-time object detectors," 2022, doi: 10.48550/ARXIV.2207.02696.
[15] C.-Y. Wang, I.-H. Yeh, and H.-Y. M. Liao, "YOLOv9: Learning What You Want to Learn Using Programmable Gradient Information," 2024, *arXiv*. doi: 10.48550/ARXIV.2402.13616.
[16] Glenn Jocher, Ayush Chaurasia, and Jing Qiu, *YOLO11*. (Sep. 10, 2024). Ultralytics. [Online]. Available: https://github.com/ultralytics/ultralytics
[17] Y. Tian, Q. Ye, and D. Doermann, "YOLOv12: Attention-Centric Real-Time Object Detectors," 2025, *arXiv*. doi: 10.48550/ARXIV.2502.12524.
[18] P. Hidayatullah, N. Syakrani, M. R. Sholahuddin, T. Gelar, and R. Tubagus, "YOLOv8 to YOLO11: A Comprehensive Architecture In-depth Comparative Review," 2025, *arXiv*. doi: 10.48550/ARXIV.2501.13400.
[19] J. Liu *et al.*, "Muon is Scalable for LLM Training," 2025, *arXiv*. doi: 10.48550/ARXIV.2502.16982.
[20] Ultralytics, "Image Classification." Accessed: Feb. 16, 2026. [Online]. Available: https://docs.ultralytics.com/tasks/classify/